# Improving Pain Classification using Spatio-Temporal Deep Learning Approaches with Facial Expressions

Aafaf Ridouan[1], Amine Bohi[1], Youssef Mourchid[1]
[1]CESI LINEACT Laboratory, UR 7527, Dijon, France

**ABSTRACT**

Pain management and severity detection are crucial for effective treatment, yet traditional self-reporting methods are subjective and may be unsuitable for non-verbal individuals (people with limited speaking skills). To address this limitation, we explore automated pain detection using facial expressions. Our study leverages deep learning techniques to improve pain assessment by analyzing facial images from the Pain Emotion Faces Database (PEMF). We propose two novel approaches[1]: (1) a hybrid ConvNeXt model combined with Long Short-Term Memory (LSTM) blocks to analyze video frames and predict pain presence, and (2) a Spatio-Temporal Graph Convolution Network (STGCN) integrated with LSTM to process landmarks from facial images for pain detection. Our work represents the first use of the PEMF dataset for binary pain classification and demonstrates the effectiveness of these models through extensive experimentation. The results highlight the potential of combining spatial and temporal features for enhanced pain detection, offering a promising advancement in objective pain assessment methodologies.

## 1. INTRODUCTION

According to the International Association for the Study of Pain (IASP), pain is defined as an unpleasant sensory and emotional experience associated with, or resembling that associated with, actual or potential tissue damage. Managing pain and detecting its severity play essential roles in treating diseases. Traditionally, this is done through self-reporting, where patients rate their pain level on a scale. Various pain measurement scales have been designed to describe a patient's intensity of pain, including the Visual Analogue Scale (VAS), Verbal Rating Scale (VRS), Faces Pain Scale-Revised (FPSR), and the Numerical Rating Scale (NRS) [1] [2]. However, these approaches depend on the patient's self-rating of pain, making them less objective, as the conceptualization of pain varies from patient to patient and between clinicians and patients [3]. Thus, relying solely on self-reporting can lead to unsuitable and inadequate pain management. In the literature, wide ranges of approaches based on facial expressions have been used for pain detection. It has been found that pain generates spontaneous facial expressions, which can be used to automate pain assessment by detecting it or estimating its level. Moreover, most publicly available pain databases include facial video frames of patients, providing valuable data for analysis. As a result, some approaches have been employed in this context, ranging from traditional methods to machine learning and deep learning approaches. Deep neural networks have shown remarkable effectiveness in automatic recognition tasks [31], including pain assessment. With the use of multiple hidden layers, these models' capacity to perform complex and highly nonlinear predictive modeling tasks has increased. Convolutional Neural Network (CNN) models achieve higher performance in image classification and feature extraction [4]. Adapting them to facial feature extraction has improved the task of pain detection based on facial image datasets. On the other hand, Recurrent Neural Networks (RNNs) are known for their ability to manage sequential data by capturing its temporal aspects [32]. By combining both CNN and RNN, we enhance the extraction of spatio-temporal features, leading to greater efficiency in classification tasks.

In this work, we propose two networks for processing two different types of data. The first is a hybrid ConvNeXt model combined with Long Short Term Memory (LSTM) [5] blocks, designed to process video frames from the Pain Emotion Faces Database (PEMF) [6] to predict the presence of pain in these videos. The second is a Spatio-Temporal Graph Convolution Network STGCN-LSTM network that uses landmarks extracted from the same dataset to predict pain. The contributions of this work are as follows:

---

[1] https://github.com/Aafaf-rido/Pain-assessment/tree/main

- Implementation of pain classification architectures using the PEMF dataset. To our knowledge, we are the first to use the facial clips from the PEMF dataset specifically for binary pain classification.
- Proposing a hybrid ConvNeXt network that combines both the ConvNeXt model that captures pain-related details in the facial expressions with LSTM blocks that handle temporal information of pain expressions across consecutive video frames.
- Implementing a Spatio-Temporal Graph Convolution Network (STGCN) with enhanced LSTM blocks using extracted landmarks from the PEMF facial frames to extend the dataset for further experiments.
- The efficiency of the two proposed architectures is shown through extensive experimentation on the PEMF dataset.

The rest of the paper is organized as follows. In section 2 we discuss the related previous research works. Section 3 represents the proposed architectures in more depth. Experiments and the results are discussed in section 4. Finally, section 5 provides a summary and discusses further the research directions.

## 2. RELATED WORK

Pain assessment is a critical component in medical diagnosis and treatment, particularly for conditions where pain is a primary symptom. Over the years, various methods have been developed to assess pain, ranging from handcrafted methods to advanced machine learning (ML) and deep learning (DL) techniques. Pikulkaew et al. [7] propose deep learning system using CNN ResNet-34 developed on the UNBC-McMaster Shoulder Pain Archive database [8]. This database consists of facial images of patients with shoulder pain to evaluate pain severity, classifying it into three levels: not painful, becoming painful, and painful. Similarly, Gholami et al. [9] performed binary pain classification on infants using a relevant vector machine (RVM) trained on the COPE dataset [10]. In the context of multimodal-based approaches, Semwal et al. [11] proposed a spatio-temporal behavioral multiparametric pain assessment system that combines three different behavioral pain parameters: facial expression, body movement, and pain sound. Three classifiers were introduced for each modality: CNNTL-BiLSTM for facial expressions, VGGish-SVM for audio, and SFCN-BiLSTM for body movements. A decision-level fusion approach was used to combine the outputs of these classifiers to estimate pain intensity as No pain, Mild pain, and Severe pain. They also introduced the Behavioral Multiparametric Pain dataset (BMP dataset) to support this approach. Bargshady et al. [12] proposed a hybrid learning-based algorithm using CNN and BiLSTM networks trained on facial images to estimate pain intensities. 3D convolutional networks have also shown promising results; in [13], a deep 3D convolutional network was used for pain detection from facial images, although these models remain computationally expensive. Y. Li et al. [14] used the EmoPain body movement dataset to predict pain recognition and protective behavior. Their lightweight LSTM-DNN model predicts pain intensity and protective behavior based on angle, angle energy, and sEMG features. Zhou et al. [15] introduced a real-time regression framework based on recurrent convolutional neural networks for automatic pain intensity estimation on the UNBC-McMaster Shoulder Pain Archive dataset. G. Bargshady et al. [16] developed an ensemble deep learning framework (EDLM) with two stages. The early fusion stage combined a fine-tuned VGG-Face with linear PCA to extract and select features from facial images. These features were then transferred to a hybrid deep learning network for classification. A three-stream ensemble CNN-RNN classifier was used in the late fusion stage to classify pain levels into five categories. Both the MIntPAIN and UNBC-McMaster Shoulder Pain datasets were used to train and evaluate the model. Bargshady et al. [17] also proposed a predictive modeling framework that employs VGG-Face with PCA using Hue, Saturation, Value (HSV) color space video frames for pain intensity estimation, followed by a modified Temporal Convolutional Network (TCN) for prediction using the UNBC-McMaster Shoulder Pain Archive and MIntPAIN databases. F. Paol et al. [18] used facial expressions to evaluate pain levels based on Action Units (AUs) and head pose components. They trained and compared multiple models using both the BioVid and the PEMF datasets. This progression highlights the advancements in pain assessment methods, emphasizing the need for future research to focus on optimizing these models for real-time applications and exploring more comprehensive multimodal datasets to improve generalizability and robustness.

## 3. METHODOLOGY

In this work, we propose two approaches to detect pain. The first approach involves a hybrid CNN-LSTM model that uses facial image data to extract both spatial and temporal features for detecting pain in image clips. The second approach implements an STGCN model to process landmarks extracted from the PEMF facial images to detect pain. The following paragraphs will detail each of these approaches.

### 3.1 Visual spatio-temporal network

In the hybrid CNN-LSTM spatio-temporal network illustrated in block a of Figure 1, twenty frames of each clip in the PEMF dataset are fed into a CNN model precisely a ConvNeXt model to handle the spatial aspect of the data.

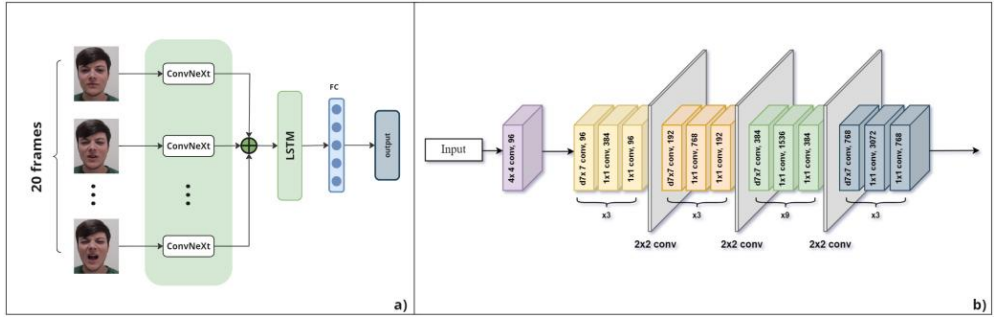

Figure 1. (a) Hybrid ConvNeXt architecture. (b) General architecture of the ConvNeXt block (Tiny version).

The ConvNeXt model [19], introduced in 2022 by Facebook researchers, is a convolutional network with design elements inspired by vision transformers [20]. It was developed by iteratively modifying ResNet [21] architectures to integrate features proven effective in vision transformers. The architecture employs multiple stages with varying feature map resolutions, focusing on stage-compute-ratio (SCR) and stem cell structure. A key feature of ConvNeXt is its patchify layer, which uses a 4×4 convolutional layer with a stride of 4, in contrast to ResNet's 7×7 layer with a stride of 2.

Figure 1. (b) illustrates the general architecture of ConvNeXt series networks. The ConvNeXt block, crucial for enhancing performance and reducing floating-point operations (FLOPs), achieves this through larger kernel-sized depthwise convolutions and a wider channel range, expanding to 96 channels. It adopts an inverted bottleneck design, uses GELU [22] activation instead of ReLU [23], and employs Layer Normalization [24] rather than BatchNormalization [25] techniques commonly found in advanced Transformers. The ConvNeXt block includes a 7×7 depthwise convolution, two 1×1 layers, and GELU activation, with Layer Normalization applied before the 1×1 convolution layer. It also incorporates a downsampling layer between stages using 2×2 convolutional layers with a stride of 2.

Multiple versions of ConvNeXt have been developed, ranging from Tiny to XLarge, with channel numbers ranging from (96, 192, 384, 768) in the Tiny version (illustrated in Figure 1. (b)) to (256, 512, 1024, 2048) in the XLarge version, and the number of blocks being (3, 3, 9, 3) in Tiny and (3, 3, 27, 3) in the larger ConvNeXt models. The extracted spatial feature vectors are then processed in a middle fusion stage. To construct the clip feature vector, the feature vectors from the twenty video frames are concatenated. These concatenated vectors are then fed into an LSTM model, which handles the temporal sequential aspects of the facial clips.

Long Short-Term Memory (LSTM) networks are an advanced type of Recurrent Neural Networks (RNNs) designed to capture temporal dependencies across sequences of data. Traditional RNNs struggle with retaining information over long sequences, as they tend to give more weight to recent information, which limits their ability to capture dependencies over a large number of timesteps. LSTM networks address this issue by introducing memory cells that can retain information over long sequences. Each memory cell contains three main components: an input gate, a forget gate, and an output gate. The input gate determines which inputs to store in the memory cell based on the current input and the previous hidden state, returning values between 0 and 1. The forget gate specifies which information to discard, with 0 meaning ignored and 1 meaning retained, based on the current input and the previous hidden state. The output gate controls how much of the memory cell's content contributes to computing the hidden state, also producing values between 0 and 1. These gating mechanisms enhance the performance of LSTM networks compared to traditional RNNs by effectively controlling the flow of information in and out of the memory cells, thus mitigating the vanishing gradient problem.

### 3.2 STGCN for landmarks data

Face landmarks represent specific key points on the face used to identify and track facial features in computer vision tasks. These key points are often connected to form a mesh representing the structure of the face. Therefore, the landmarks with their connections construct a graph, where each landmark represents a node in the graph and is connected to other landmarks via edges. As a contribution to this work, we extended the dataset by extracting the landmarks from every facial video clip using the dlib library. Dlib [26] provides robust facial landmark detection through its pre-trained models. We looped through the frames of each of the 272 clips (videos) to extract 68 landmarks representing essential key points in facial images. Each of these key points is represented by 2-dimensional coordinates (x, y) and saved in a CSV file, associating each image with its 68 landmark coordinates. The pipeline of landmarks extraction using Dlib library is represented in Figure 3.

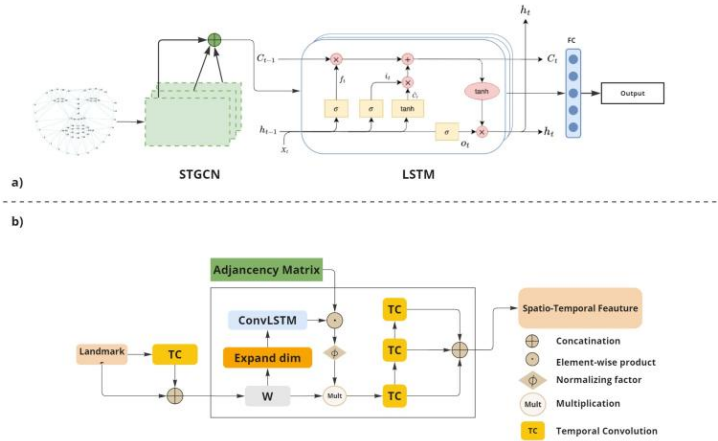

Figure 2. (a) STGCN-LSTM framework overview. (b) STGCN block.

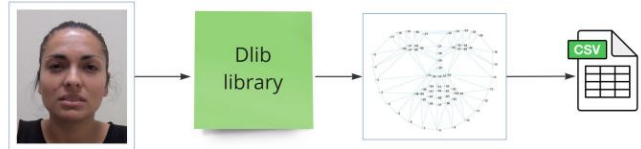

Figure 3. Landmarks extraction pipeline using Dlib library.

Recently, graphs have been widely used in various applications [33, 34, 35]. Graph Convolutional Networks (GCNs) were introduced to handle the graph-structured data. Unlike other convolution networks, GCNs combine the information of the node itself, in our case the landmarks, with the information from its neighboring nodes through the edges. This mechanism allows this type of network to capture both local and global patterns in the data since it updates the node features at every step by combining them with the features of their neighbors. Since our input in the second model is landmarks extracted from video frames, it is important to handle both the spatial and temporal aspects of it. Therefore, we implemented a Spatial-Temporal Graph Convolutional Networks STGCN inspired by the work represented in [27]. STGCNs are an advanced version of GCNs. They extend the capabilities of GCNs by incorporating both spatial and temporal dimensions by introducing a temporal convolution block. While GCNs focus on capturing spatial relationships within a static graph, STGCNs are designed to handle dynamic graph data that evolves over time, making them suitable for our task.

Figure 2. (b) represents an STGCN block, which mainly consists of Temporal Convolutions (TC) that compute the temporal dependencies within the data, and a ConvLSTM layer for combining spatial and temporal information. It also uses an adjacency matrix that defines the connections between facial landmarks, helping to construct a graph structure. Combining these components with some additional operations such as concatenation, element-wise product,

normalization, and multiplication, which help refine the features, all these components work together to produce spatio-temporal features. These features are then fed into stacked LSTM blocks with two layers to emphasize handling the temporal aspect of the sequential landmark data. A fully connected layer is used at the end for the pain classification task. This process is illustrated in Figure 2. (a) Combining STGCN and LSTM improves pain classification by capturing both the spatial relationships between facial landmarks and the changes in their movements over time. STGCN is good at representing spatial features using adjacency matrices and convolution operations, while LSTM excels at understanding long-term changes in the data. By working together, the model uses STGCN for detailed spatial structure and LSTM for recognizing temporal patterns, resulting in more accurate pain classification. This combination helps the model detect small changes in facial expressions, which is important for assessing pain effectively.

## 4. EXPERIMENT AND RESULTS

In this work, we propose two approaches to detect pain. The first approach involves a hybrid CNN-LSTM model that uses facial image data to extract both spatial and temporal features for detecting pain in image clips. The second approach implements an STGCN model to process landmarks extracted from the PEMF facial images to detect pain. The following paragraphs will detail each of these approaches.

### 4.1 Dataset

The experiments were conducted using a recent dataset called the Pain Emotion Faces (PEMF) database [6]. The PEMF dataset comprises 272 micro-clips collected from 68 healthy participants with an average age of 30.34 years. It includes two main classes: "No Pain" (neutral) and three types of pain-related facial expressions: posed pain, spontaneous pain (measured with an algometer), and spontaneous pain (induced by $CO_2$ laser). The dataset was evaluated by 510 undergraduate students who rated pain intensity, valence, arousal, and the presence of additional emotions (happiness, sadness, anger, surprise, fear, and disgust). Each clip consists of 20 frames, available in both color and black-and-white versions. Additional metrics include pain intensity, the percentage of six emotions, and facial action units (AUs). In this work, we use the frames as input to the hybrid ConvNeXt model and generate new landmark data from them for use with the STGCN-LSTM network.

### 4.2 Experiment 1: Visual spatio-temporal network

#### 4.2.1 Implementation details

In our training process, we used the Adam optimizer with a learning rate of 1e-4, combined with exponential decay to ensure the convergence of the hybrid ConvNeXt model. Data augmentation techniques, such as Random Cropping and Random Rotation, were applied to improve the model's performance and generalization capabilities. Additionally, to further enhance our model's performance, we included weights from a ConvNeXt model pre-trained on the ImageNet-22k dataset [28]. This dataset, which includes a diverse range of images, helps our model learn more effectively. We resized our training images to 224×224 pixels to match the input size used for the pre-trained weights, facilitating their effective utilization and contributing to improved performance and accuracy.

In our study, we chose the XLarge version of the ConvNeXt model, based on the scores obtained for image classification on the ImageNet dataset [21] and the study [29] where the authors conducted extensive experiments on the five versions of the ConvNeXt model for facial emotion classification, namely: Tiny, Small, Base, Large, and XLarge. The ConvNeXt XLarge demonstrated superior performance. For sequence learning, we implemented four types of LSTM: the original LSTM, Bi-LSTM, LSTM with an integrated attention mechanism, and stacked LSTM. The objective is to test all these LSTM variants and select the one that provides the best performance in combination with the ConvNeXt XLarge. For more information on the different LSTM versions, interested readers can refer to this study [30]. Finally, the model was trained for 150 epochs with a batch size of 8. We used 80% of the data for training, while the remaining 20% was split between the test and validation sets. The results are represented in the following section.

#### 4.2.2 Results

To identify the most effective CNN model for our pain classification task, we evaluated several pre-trained CNN models on the PEMF dataset, selecting the best-performing ones for detailed comparison. As presented in Table 1, the ConvNeXt XLarge model achieved the highest accuracy of 89.01%, significantly outperforming the other models. VGG16 and Xception followed with accuracies of 87.39% and 85.59% respectively. In contrast, models such as ResNet101, InceptionV3, and MobileNetV1 demonstrated lower performance, with accuracies of 78.92%, 68.9%, and 66.11% respectively. These results highlight the superior ability of ConvNeXt XLarge to capture spatial features crucial for pain classification based on facial images. Therefore, we have chosen the ConvNeXt XLarge model as the spatial feature extractor for our network.

To optimize the sequence learning component of our network, we implemented and evaluated different variants of the LSTM model. The results, presented in Table 2, reveal the average performance of each variant across five experiments (5-folds) in terms of accuracy, precision, recall, and F1-score. The LSTM with an attention mechanism achieved a moderate accuracy of 86.67%, indicating that while the attention mechanism adds value, it did not outperform the other variants in this specific task. The standard LSTM and Bi-LSTM versions demonstrated robust performance, with accuracies surpassing 90%. However, the stacked LSTM variant outshined all other models, achieving the highest accuracy of 93.51%, along with superior precision, recall, and F1-score metrics. These results indicate that the ConvNeXtXL-Stacked LSTM combination is the most effective for pain classification, effectively capturing both spatial and temporal features in the data.

Table 1: Comparison of CNN models on the PEMF dataset, showing accuracy scores

| Model | Accuracy (%) |
|---|---|
| MobileNetV1 | 66.11 |
| InceptionV3 | 68.9 |
| Resnet101 | 78.92 |
| Xception | 85.59 |
| VGG16 | 87.39 |
| **ConvNeXtXL** | **89,01** |

Table 2: Performance comparison of ConvNeXt XLarge with different LSTM variants

| Model | Accuracy (%) | Precision (%) | Recall (%) | F1-score (%) |
|---|---|---|---|---|
| ConvNeXtXL - LSTM | 90.45 | 91.02 | 90.45 | 90.54 |
| ConvNeXtXL - Bi-LSTM | 90.27 | 90.21 | 90.27 | 90.13 |
| ConvNeXtXL - LSTM with attention | 86.67 | 87.23 | 86.67 | 86.70 |
| **ConvNeXtXL - Stacked LSTM** | **93,51** | **93.97** | **93,51** | **93.61** |

## 4.3 Experiment 2: STGCN for landmarks data

### 4.3.1 Training details

The STGCN-LSTM is trained on the landmarks data using the Adam optimizer implemented with Exponential decay and a learning rate of the value 1e-4. To prevent overfitting, the SMOTE method (Synthetic Minority Over-sampling Technique) was implemented to address the class imbalance in the dataset. It works by creating synthetic examples of the minority class to balance the number of instances between the majority and minority classes. The model converges at 150 epochs with a batch size of 10. In the next section, we present the results of the training.

### 4.3.2 Results

To ensure the effectiveness of the STGCN model on the pain classification task using landmarks data, we ran the model five times on the training and testing datasets. We started by implementing an STGCN model with three blocks and then combined it with an LSTM with two layers. The results in Table 3 show a significant difference between the STGCN with and without the LSTM model. With an accuracy reaching 82.14%, the combination of the STGCN with the LSTM improved the results, emphasizing the temporal aspect of the landmarks data. Thus, relying solely on the STGCN is not efficient for the pain classification task based on clips of landmarks.

Table 3: STGCN models results on landmarks dataset

| Model | Accuracy (%) | Precision (%) | Recall (%) | F1-score (%) |
|---|---|---|---|---|
| STGCN | 78,93 | 78 | 78,80 | 78,40 |
| **STGCN-LSTM** | **82,14** | **81,60** | **82** | **81,80** |

## 5. CONCLUSION

This paper proposes two different approaches built on the same dataset PEMF which is considered new with very few applications on the pain assessment task. Using the micro-clips (videos) of the dataset, we implemented a hybrid ConvNeXt model that handles both the spatial and temporal aspects of the facial clip frames. The pre-trained ConvNeXt XLarge model extracted the spatial features of each clip frame passing them to an LSTM model that studied the sequential dependencies between the video frames. The hybrid network shows significant performance on pain classification task.

The second approach is based on the extracted landmarks from the PEMF clip frames. These landmarks are key points connected to form a graph with nodes (landmark coordinates) and edges (connections). To evaluate pain using the landmarks graph data, we implemented an STGCN-LSTM model that studies the connections between the landmarks through the video frames. Experimental results of the STGCN-LSTM model on the constructed PEMF landmark dataset demonstrate good performance for the pain classification task. The integration of the LSTM improved the results, providing more accurate facial pain detection.

In future work, we intend to combine the two proposed approaches to build a multimodal network, combining the hybrid ConvNeXt model with the STGCN-LSTM and using the multimodal data facial image landmarks PEMF dataset. Given the results of the two approaches separately, we count on building a strong multimodal pain detection framework combining the two networks.